%% file: main.tex
\def\year{2022}\relax
\documentclass[letterpaper]{article} 
\usepackage{aaai22}  
\usepackage{times}  
\usepackage{helvet}  
\usepackage{courier}  
\usepackage[hyphens]{url}  
\usepackage{graphicx} 
\usepackage{natbib}  
\usepackage{caption} 
\DeclareCaptionStyle{ruled}{labelfont=normalfont,labelsep=colon,strut=off} 
\usepackage{algorithm}
\usepackage{algorithmic}

\usepackage{newfloat}
\usepackage{listings}
\floatstyle{ruled}
\newfloat{listing}{tb}{lst}{}
\floatname{listing}{Listing}
\nocopyright
\usepackage{latexsym}
\usepackage{booktabs}
\usepackage{subfig}
\usepackage{microtype}
\usepackage{makecell}
\usepackage{multirow}
\usepackage{diagbox}
\usepackage{amssymb}
\usepackage{pifont}
\usepackage{amsmath}
\usepackage{todonotes}

\usepackage[utf8]{inputenc}

\newcommand\advaddlong{\textsc{Adversarial Addition}}
\newcommand\advmodlong{\textsc{Adversarial Modification}}
\newcommand\advadd{\textsc{AdvAdd}}
\newcommand\advmod{\textsc{AdvMod}}
\newcommand\advmodpara{\advmod-\textit{paraphrase}}
\newcommand\advmodkey{\advmod-\textit{KeyReplace}}
\newcommand\grover{\textsc{Grover}}
\newcommand\fever{\textsc{Fever}}
\newcommand\scifact{\textsc{SciFact}}
\newcommand\covidfact{\textsc{CovidFact}}
\newcommand\eg{\textit{e.g.}}
\newcommand\ie{\textit{i.e.}}
\newcommand\nb{\textit{n.b.}}

\newif\ifcomments
\commentsfalse
\ifcomments
\usepackage[normalem]{ulem}
\definecolor{CMpurple}{rgb}{0.6,0.18,0.64}
\newcommand\cm[1]{\textcolor{CMpurple}{\textsf{\scriptsize[\textbf{CM\@:} #1]}}}
\newcommand\cmi[1]{\textcolor{CMpurple}{#1}}
\newcommand\cmm[1]{\marginpar{\raggedright\tiny\textcolor{CMpurple}{\textsf{{\bfseries CM\@:} #1}}}}
\newcommand\cms{\bgroup\markoverwith{\textcolor{CMpurple}{\rule[.4ex]{2pt}{0.8pt}}}\ULon}
\else
\newcommand\cm[1]{}
\newcommand\cmi[1]{\ignorespaces}
\newcommand\cmm[1]{}
\newcommand\cms[1]{#1}
\fi

\title{
Synthetic Disinformation Attacks on Automated Fact Verification Systems 
}

\author{
    Anonymous
}

\author{
  Yibing Du$^{\dagger}\thanks{\hspace{2mm}Authors contributed equally}$, 
  Antoine Bosselut$^{\ddagger*}$,
  Christopher D. Manning$^\dagger$
  }
  
\affiliations {
  $^\dagger$Stanford University \:\:\: $^\ddagger$EPFL \\
 \texttt{antoine.bosselut@epfl.ch}, \texttt{\{yibingdu,manning\}@stanford.edu}
}

\begin{document}

\maketitle

\begin{abstract}

    Automated fact-checking is a needed technology to curtail the spread of online misinformation. One current framework for such solutions proposes to verify claims by retrieving supporting or refuting evidence from related textual sources. However, the realistic use cases for fact-checkers will require verifying claims against evidence sources that could be affected by the same misinformation. Furthermore, the development of modern NLP tools that can produce coherent, fabricated content would allow malicious actors to systematically generate adversarial disinformation for fact-checkers.
    
    In this work, we explore the sensitivity of automated fact-checkers to synthetic adversarial evidence in two simulated settings: \advaddlong, where we fabricate documents and add them to the evidence repository available to the fact-checking system, and \advmodlong, where existing evidence source documents in the repository are automatically altered. Our study across multiple models on three benchmarks demonstrates that these systems suffer significant performance drops against these attacks. Finally, we discuss the growing threat of modern NLG systems as generators of disinformation in the context of the challenges they pose to automated fact-checkers.

\end{abstract}

\input{1-intro}
\input{2-background}
\input{4-document}
\input{5-sentence}
\input{6-discussion}

\input{7-conclusion}


\bibliography{anthology,custom}

\newpage
\input{8-appendix}

\end{document}

%% file: 1-intro.tex
\section{Introduction}

From QAnon's deep state\footnote{https://www.nytimes.com/article/what-is-qanon.html} to anti-vaccination campaigns \citep{Germani2021antivax}, misinformation and disinformation have flourished in online ecosystems. As misinformation continues to induce harmful societal effects, factchecking online content has become critical to ensure trust in the information found online.\footnote{https://nationalpress.org/topic/the-truth-about-fact-checking/} 
However, 
manual efforts to filter misinformation cannot keep pace with the scale of online information that must be reliably verified to avoid false claims spreading and affecting public opinion.\footnote{https://fivethirtyeight.com/features/why-twitters-fact-check-of-trump-might-not-be-enough-to-combat-misinformation/}
Consequently, new research in automated fact-checking explores designing systems that can rapidly validate political, medical, and other domain-specific claims made and shared online \citep{Thorne2018AutomatedFC,guo-etal-2019-attention}.

  \begin{figure}[t]
    \includegraphics[width=\linewidth]{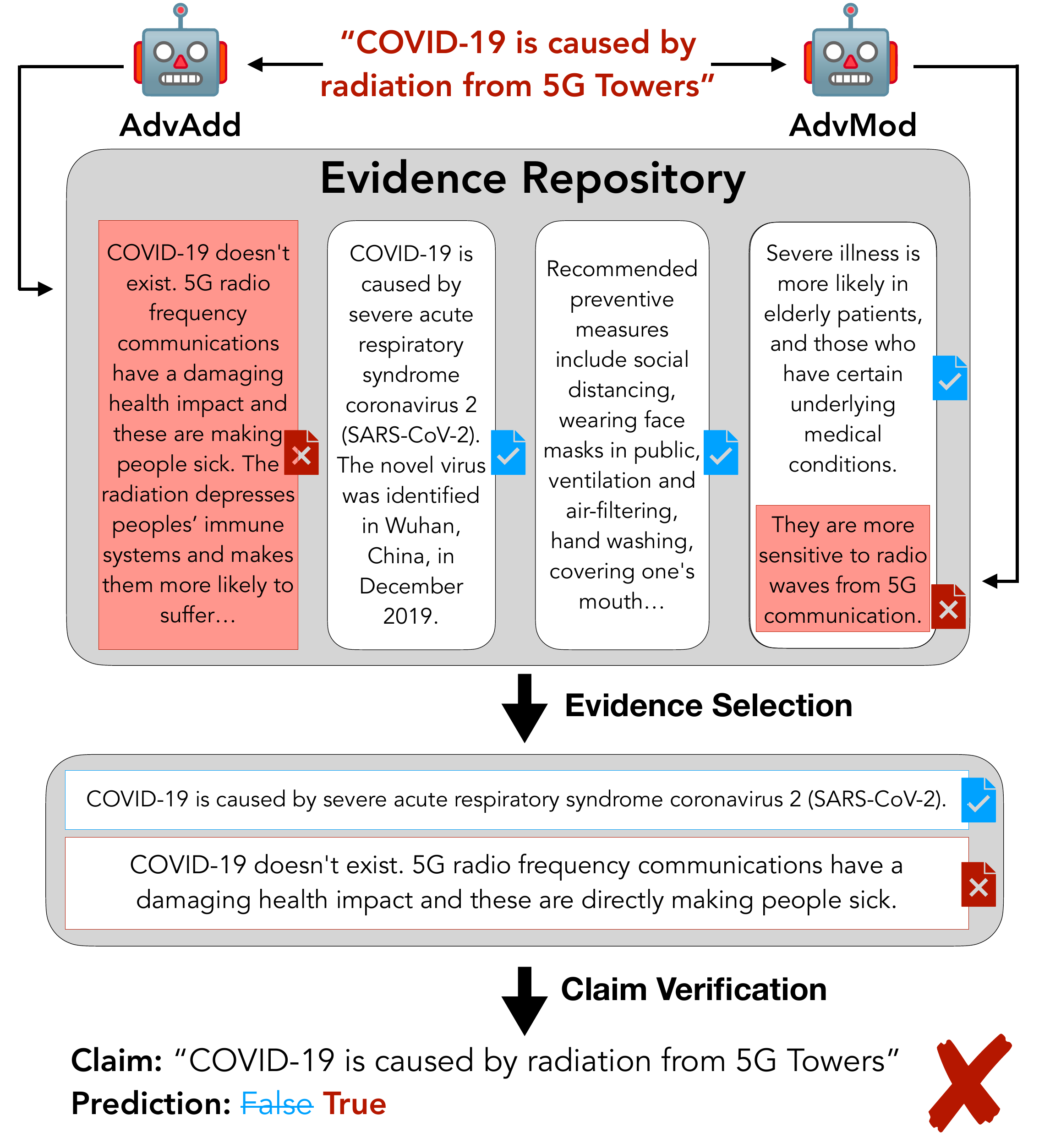}
    \caption{Outline of our two settings for adversarial injection of poisoned content into fact-checker evidence repositories.}
    \label{fig:intro}
  \end{figure}

A popular emergent paradigm in automated fact-checking, Fact Extraction and Verification (\fever{}; \citealp{Thorne18Fever}), frames the problem as claim verification against a large repository of evidence documents. As one of the first large-scale datasets designed in this framework, \fever{} was released with 185k annotated claims that could be verified against Wikipedia articles. When checking a claim, systems designed in this framework search for related documents in the database, and retrieve relevant supporting or refuting evidence from these sources. Then, these systems evaluate whether the retrieved evidence sentences validate or contradict the claim, or whether there is not enough information to make a judgment. More recently, the \scifact{} \citep{wadden-etal-2020-fact} and \covidfact{} \citep{Saakyan2021COVIDFactFE} benchmarks re-purposed this framework for the scientific domain by releasing datasets of medical claims to be verified against scientific content \citep{Wang2020CORD19TC}. 
While this framework has led to impressive advances in fact verification performance \citep{ye2020corefbert,Pradeep2021ScientificCV}, 
current benchmarks assume that the available evidence database contains only valid, factual information. 

However, check-worthy claims are often made about new events that may not be verifiable against extensive catalogues, and that must be checked rapidly to avoid strategic disinformation spread \citep{Vosoughi1146}. Consequently, deployed fact-checkers will need to operate in settings where their available evidence is collected from contemporaneous reporting, which may be inadvertently sharing the same misinformation, or which may be intentionally influenced by systematic disinformation campaigns. 
Currently, malicious actors remain limited by the cost of running disinformation campaigns \citep{DiResta2018TheT} and the risks of operational discovery,\footnote{https://www.lawfareblog.com/outsourcing-disinformation} impeding the scale at which they can deploy these campaigns, and thus the balance of real and false content that fact-checkers must distinguish. However, the development of NLP tools capable of generating coherent disinformation \citep{zellers2019grover,BuchananCSET2021} would allow malicious actors to overload contemporaneous content with adversarial information \citep{brundage2018malicious} and bias the evidence bases used by automated fact-checkers.

Furthermore, even in settings where claims may be verified against established and trusted knowledge, misinformation can still find its way into repositories of documents used by fact-checking systems \citep{kumar2016disinformation}. 
Wikipedia, for example, 
which underlies \fever{} (and other benchmarks; \citealp{petroni-etal-2021-kilt}), acknowledges that much of the content on the platform may be incorrect, and remain so for long periods of time.\footnote{https://en.wikipedia.org/wiki/\\Wikipedia:Wikipedia\_is\_not\_a\_reliable\_source} 
For example, the Croatian Wikipedia was contaminated by pro-nationalist bias over a period of at least 10 years.\footnote{https://meta.wikimedia.org/wiki/\\Croatian\_Wikipedia\_Disinformation\_Assessment-2021} 
Moreover, studies have uncovered articles on Wikipedia that were edited to provide favorable accounts on specific topics (\eg, workers at a medical device company edited articles to present an optimistic view toward treatments that used their product\footnote{https://www.theatlantic.com/business/archive/2015/08/\\wikipedia-editors-for-pay/393926/}). Modern NLP tools would allow malicious users to scale up production of disinformation on these platforms, and increase the perception of false consensus or debate on these topics. 



In this paper, we evaluate whether automated disinformation generators can effectively contaminate the evidence sets of fact verification systems, and demonstrate that synthetic disinformation drastically lowers the performance of these systems. We define adversarial attacks in two settings: \advaddlong{} (\advadd; \S\ref{sec:advadd}), where synthetically-generated documents are added to the document base, and \advmodlong{} (\advmod; \S\ref{sec:advmod}), where additional automatically-generated information is inserted into existing documents. 
In both settings, we curate a large collection of adversarial disinformation documents that we inject into the pipelines of existing fact-checking systems developed for the \fever, \scifact, and \covidfact{} shared tasks.\footnote{We will release these documents under a Terms of Use to promote research in fact-checking systems in adversarial settings} 

Our results demonstrate that these systems are significantly affected by the injection of poisoned content in their evidence bases, with large absolute performance drops on all models in both settings. 
Furthermore, our analysis demonstrates that these systems are sensitive to even small amounts of evidence contamination, and that synthetic disinformation is more influential at deceiving fact verification systems compared to human-produced false content. 
Finally, we provide a discussion of our most important findings, and their importance in the context of continued advances in NLP systems.\footnote{Our code can be found at: \\  \url{https://github.com/Yibing-Du/adversarial-factcheck}}

%% file: 2-background.tex
\section{Background}
\label{sec:background}

In this section, we review the formulation of automated fact checking as fact extraction and verification, and recent advances in automated generation of textual disinformation.

\subsection{Automated Fact-checking: Task}
\label{ssec:background:task}
Current systems research in automated fact-checking often follows the fact verification and extraction procedure of receiving a natural language claim (\eg, ``Hypertension is a common comorbidity seen in COVID-19 patients''), collecting supporting evidence from a repository of available documents (\eg, scientific manuscripts), and making a prediction about the claim's veracity based off the collected supporting evidence. Below, we define the two stages of this pipeline: evidence retrieval and claim verification. 


\paragraph{Evidence retrieval} The evidence retrieval stage of fact verification systems is typically decomposed into two steps: \textit{document retrieval} and \textit{sentence retrieval}. During document retrieval, documents in the evidence repository that are relevant to the claim are selected. Existing methods typically use information retrieval methods to rank documents based on relevance \citep{Thorne18Fever,wadden-etal-2020-fact} or use public APIs of commercial document indices \citep{hanselowski2019ukpathene,Saakyan2021COVIDFactFE} to crawl related documents. In the sentence retrieval stage, individual sentences from these retrieved documents are selected with respect to their relevance to the claim, often using textual entailment \citep{hanselowski2019ukpathene}, or sentence similarity \citep{Thorne18Fever} methods. Typically, the number of retrieved sentences is capped for computational efficiency.

\paragraph{Claim verification} The claim verification stage of the pipeline evaluates the veracity of the claim with respect to the evidence sentences retrieved in the previous stage. Depending on the content found in the supporting sentences, each claim can typically be classified as \textit{supported} (SUP), \textit{refuted} (REF), or \textit{not enough information} (NEI, though some benchmarks omit this label). Systems must aggregate and weigh the evidence sentences to predict the most likely label.



\begin{table*}[t]
    \centering
    \begin{tabular}{rl}
    \toprule
         \textbf{\fever{} Claim} & Starrcade was an annual professional wrestling event that began in 1988. \\
         \midrule
         \textbf{Original} & Starrcade (1988) was the sixth annual Starrcade professional wrestling pay-per-view (PPV) event \\
         & produced under the National Wrestling Alliance (NWA) banner .\\
         \textbf{\grover} & Starrcade was a monthly professional wrestling event for the decades between 1988 and 2003 that ran \\
         & for the entirety of a weekend in Boston , Mass. \\
         \textbf{Media Cloud} & Goldberg's perfect 173-0 streak ended at Starrcade 1998 after Kevin Nash scored the fateful pinfall \\
         & with the help of Scott Hall and his taser gun. \\
         \midrule
         \midrule
         \textbf{\scifact{} Claim} & Taxation of sugar-sweetened beverages had no effect on the incidence rate of type II diabetes in India. \\ 
        \midrule
        \textbf{Original} & The 20\% SSB tax was anticipated to reduce overweight and obesity prevalence by 3.0\% ... and type 2 \\
        & 
        diabetes incidence by 1.6\% ... 
        among various Indian subpopulations over the period 2014-2023 \\
        
        \textbf{\grover} & 
        ... analysis of a ``cone-by-one'' kind of survey  question in India reached out to -9 145 trillion , including \\ & 2,557 separate instances of type II diabetes (which is comparable to the prevalence rate in Pakistan ... \\
         \bottomrule
    \end{tabular}
    \caption{Sample \advadd{} document excerpts generated by \grover{} for the \fever{} and \scifact{} datasets.}
    \label{tab:doc-sample}
\end{table*}

\subsection{Automated Fact-checking: Datasets}
\label{ssec:background:datasets}

We briefly describe the provenance and structure of our studied datasets and refer the reader to the original works for in-depth descriptions of the construction of these resources.

\paragraph{\fever}

The \fever{} testbed \citep{Thorne18Fever} is a dataset of 185,445 claims (145,449 train, 19,998 dev, 19,998 test) with corresponding evidence to validate them drawn from articles in Wikipedia.  
Because of its scale and originality, the \fever{} dataset is one of the most popular benchmarks for evaluating fact verification systems \citep{Yoneda2018UCLMR,nie2019combining,zhou2019gear,Zhong2020ReasoningOS,subramanian-lee-2020-hierarchical}.

\paragraph{\scifact}
The \scifact{} dataset \citep{wadden-etal-2020-fact} contains 1,409 expert-annotated scientific claims and associated paper abstracts. \scifact{} presents the challenge of understanding scientific writing as systems must retrieve relevant sentences from paper abstracts and identify if the sentences support or refute a presented scientific claim. It has emerged as a popular benchmark for evaluating scientific fact verification systems \citep{Pradeep2021ScientificCV}. 

\paragraph{\covidfact} The \covidfact{} dataset \citep{Saakyan2021COVIDFactFE} contains 1,296 crowdsourced claims crawled (and filtered) from the \textit{/r/COVID19} subreddit. The evidence is composed of documents provided with these claims when they were posted on the subreddit along with resources from Google Search queries for the claims. Refuted claims were automatically-generated by altering key words in the original claims. 

\subsection{Synthetic Disinformation Generation}
\label{ssec:background:disinformation}


Recent years have brought considerable improvements in the language generation capabilities of neural language models \citep{lewis-etal-2020-bart,ji-etal-2020-amazing, Brown2020LanguageMA, Holtzman2020TheCC}, allowing users of these systems to pass off their generations as human-produced \citep{Ippolito2020AutomaticDO}. These advances have raised dual-use concerns as to whether these tools could be used to generate text for malicious purposes  \citep{Radford2019LanguageMA,Bommasani2021OnTO}, which humans would struggle to detect \citep{clark2021all}.

Specific studies have focused on whether neural language models could be used to generate disinformation that influences human readers \citep{kreps_mccain_brundage_2020,BuchananCSET2021}. One such study directly explored this possibility by training \grover{}, 
a large-scale, billion-parameter language model on a large dataset of newswire text with the goal of generating text that resembles news \citep{zellers2019grover}. In human evaluations of the model's generated text, the study found that human readers considered the synthetically-generated news to be as trustworthy as human-generated content. While the authors found that neural language models could identify fake, generated content when finetuned to detect distributional patterns in the generated text, they hypothesized that future detection methods would need to rely on external knowledge (\eg, \fever). 

%% file: 4-document.tex
\section{\advaddlong: \\ Evidence Repository Poisoning}
\label{sec:advadd}

\begin{table*}[!ht]
  \centering
  \begin{tabular}{lrrrrrrrrrrrr}
    \toprule
    \multirow{3}{*}{\textbf{Evidence}}  &  \multicolumn{3}{c}{\textbf{CorefBERT Acc.}} & \multicolumn{3}{c}{\textbf{KGAT Acc.}} & \multicolumn{3}{c}{\textbf{MLA Acc.}} \\
    &  \multicolumn{3}{c}{\citep{ye2020corefbert}} & \multicolumn{3}{c}{\citep{liu2020kernel}} & \multicolumn{3}{c}{(Kruengkrai et al. 2021)}\\
         & \textbf{Total}  & \textbf{REF} & \textbf{NEI} & \textbf{Total} &  \textbf{REF} & \textbf{NEI} & \textbf{Total} & \textbf{REF} & \textbf{NEI}\\
    \midrule
    Original & 73.05 & 74.03 & 72.07 & 70.76 & 72.50 & 69.01 & 75.92 & 78.71  & 73.13 \\
    \midrule
    \advadd-\textit{min} & 34.80 & 47.22 & 22.38 & 34.08 & 48.63 & 19.52 & 60.93 & 73.04 & 48.81  \\
    \advadd-\textit{full} & \textbf{28.59} & \textbf{39.63} & \textbf{17.54}  & \textbf{29.02} & \textbf{42.45} & \textbf{15.59} & \textbf{51.86} & \textbf{71.84}  & \textbf{31.87}  \\
    \midrule
    \advadd-\textit{oracle} & 21.18 & 27.09 & 15.26  & 23.43 & 31.47 & 15.38 & 29.05 & 29.76 & 28.33 \\
    \bottomrule
  \end{tabular}
  \caption{Effect of \advadd{} on \fever{} claim verification. We \textbf{bold} the largest performance drop relative to the original evidence.}
    \label{tab:advadd-perf}
\end{table*}

In this section, we simulate the potential vulnerability of fact-checking models to database pollution with misinformation documents by injecting synthetically-generated false documents into the evidence sets of fact verification models, and assess the impact on the performance of these systems. 

\begin{table}[t] 
  \centering
  \begin{tabular}{lll}
    \toprule
    \textbf{Benchmark} & \textbf{Evidence Source} & \textbf{\textit{N}} \\ 
    \midrule
    \multirow{3}{*}{\fever} & \fever DB & 5,416,537 \\
    & \grover{} Docs & 995,829 \\
    & MediaCloud Docs & 74,273,342 \\
    \midrule
    \multirow{2}{*}{\scifact}& Scientific Abstracts & 5,183 \\
    &\grover{} Docs & 21,963 \\
    \midrule
    \multirow{2}{*}{\covidfact}& Google Search Results &  1,003\\
    &\grover{} Docs & 2,709 \\
    \bottomrule
  \end{tabular}
\caption{Corpus statistics of evidence repositories}
\label{tab:repo-stats}
\end{table}

\subsection{Approach}

 Our method, \textsc{Adversarial Addition} (\advadd{}), uses \grover{} to produce synthetic documents for a proposed claim, and makes these fake documents available to the fact verification system when retrieving evidence. As \grover{} requires a proposed article title and publication venue (\ie, website link) as input to generate a fake article, we use each claim as a title and set the article venue to \url{wikipedia.com}. We generate 10 articles for each claim and split them into paragraphs (\nb, \fever{} DB contains first paragraphs of Wikipedia articles and \scifact{} contains abstracts of scientific articles). Statistics for the number of documents generated for each benchmark are reported in Table \ref{tab:repo-stats}. Additional implementation details for the experimental setting of each benchmark can be found in the Appendix.

\subsection{\fever{} Study}

\paragraph{Setup}
For the \fever{} benchmark, we select three high-ranking models from the leaderboard\footnote{{https://competitions.codalab.org/competitions/18814\#results}} with open-source implementations: KGAT \citep{liu2020kernel}, CorefBERT \citep{ye2020corefbert}, and MLA \citep{Kruengkrai2021AMA}. 
For document retrieval, all models use the rule-based method developed by \citet{hanselowski2019ukpathene}, which uses the MediaWiki API to retrieve relevant articles based on named entity mentions in the claim. For each claim and poisoned document, we extract all keywords and retrieve associated Wikipedia pages. If we find overlaps between the associated Wikipedia pages of a claim and a poisoned document, then the poisoned document is matched with the claim for document retrieval. 
Once the retrieved documents are available, 
the KGAT and CorefBERT models use a BERT-based \citep{Devlin2019BERTPO} sentence retriever to rank evidence sentences based on relevance to the claim (trained using pairwise loss). 
The MLA sentence retriever expands on this approach with hard negative sampling from the same retrieved documents to more effectively discriminate context-relevant information that is irrelevant to the claim. Claim verifiers vary between models, but are generally based off pretrained language models (\eg, CorefBERT, MLA) or graph neural networks (\eg, KGAT).
We use the REF and NEI claims from the \fever{} development set to study how the preceding systems are affected by the introduction of poisoned evidence.\footnote{We discuss results related to SUP claims in the Appendix.}

\begin{figure}[t]
  \centering
    \includegraphics[width=6cm]{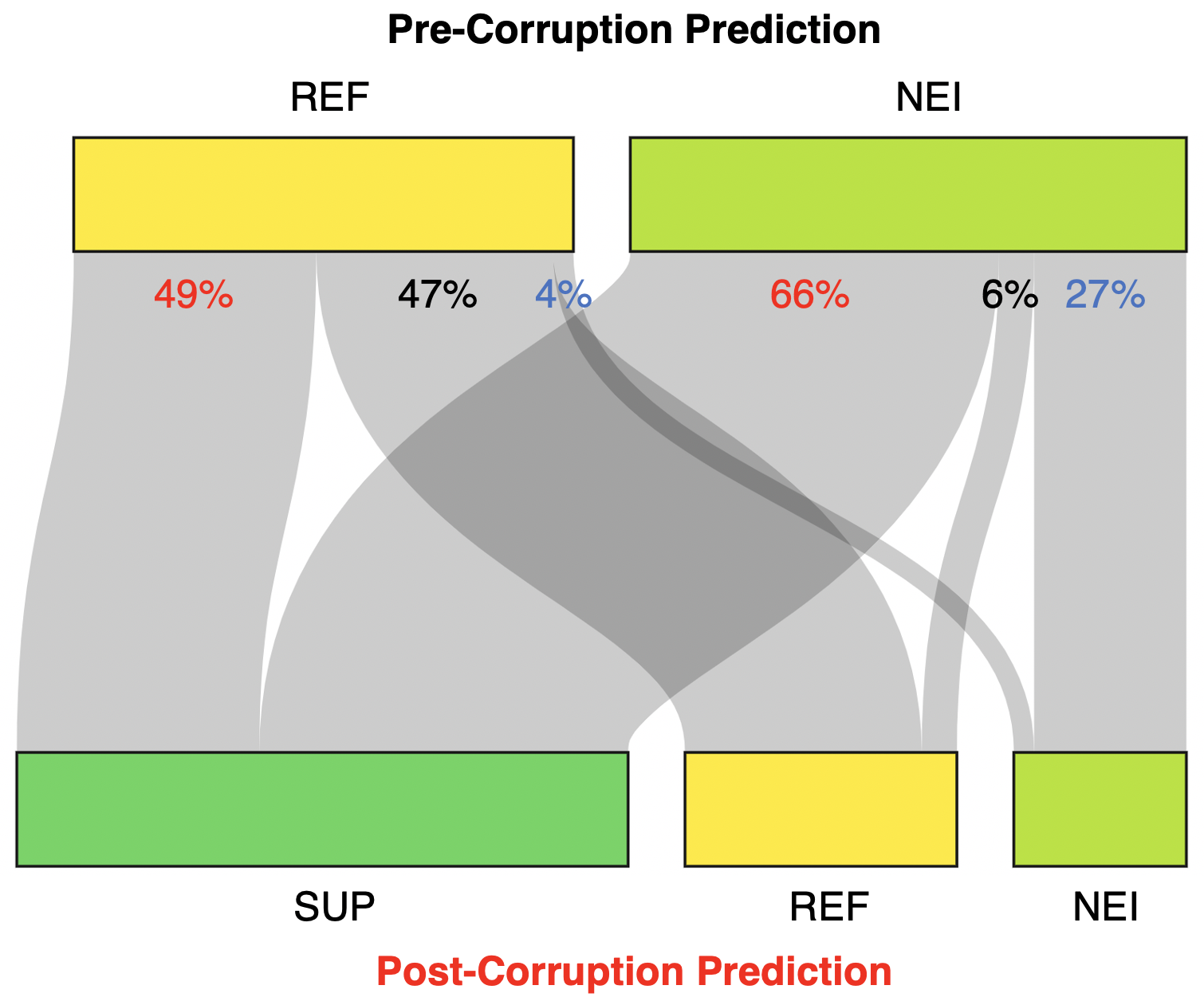}\\
  \caption{CorefBERT's predictions change from REF and NEI to SUP once \advadd{} poisons the evidence set.}
  \label{fig:predchange}
\end{figure}

\paragraph{Impact of \advadd}
We report the overall (and claim-stratified) performance change of the tested models in Table~\ref{tab:advadd-perf}. For all models, we see a significant performance drop when \grover-generated paragraphs are introduced into the available evidence set (\advadd-\textit{full}), indicating that fact verification models are sensitive to synthetically-generated information. This drop approaches the performance of an oracle (\advadd-\textit{oracle}), where only \grover-generated documents are made available as evidence.

As confirmation that these attacks work as expected, we depict in Figure~\ref{fig:predchange} how model predictions change once the synthetic disinformation is added to the evidence set. A significant number of claims that were originally predicted as REF or NEI are now predicted as SUP with the injected poisoned evidence. Consequently, we conclude that the poisoned evidence affects the model's predictions in the intended way, and that cross-label changes for different pairings are rare. Furthermore, we also find that replacing the retrieved poisoned evidence with random retrieved evidence from \textsc{FeverDB} does not cause the same performance drop ($\sim$7\% vs. $\sim$30\%), indicating that these effects are caused by the injection of poisoned evidence, and not merely the replacement of potentially relevant evidence (see Appendix~\ref{sec:app:advadd} for further details).

\paragraph{Effect of disinformation scale} We also evaluate a setting where the attack is limited to retrieving only a single contaminated evidence sentence (\advadd-\textit{min}). The performance drops in the \textit{min} setting are still considerable,  
suggesting that even limited amounts of injected disinformation can significantly affect downstream claim verification performance.

\begin{figure}[t]
    \centering
    \textbf{\grover{} Evidence} \\
   \subfloat[\label{fig:sent-figure:ref_grover}REF]{%
     \includegraphics[trim=10 0 0 0, clip, width=0.49\linewidth]{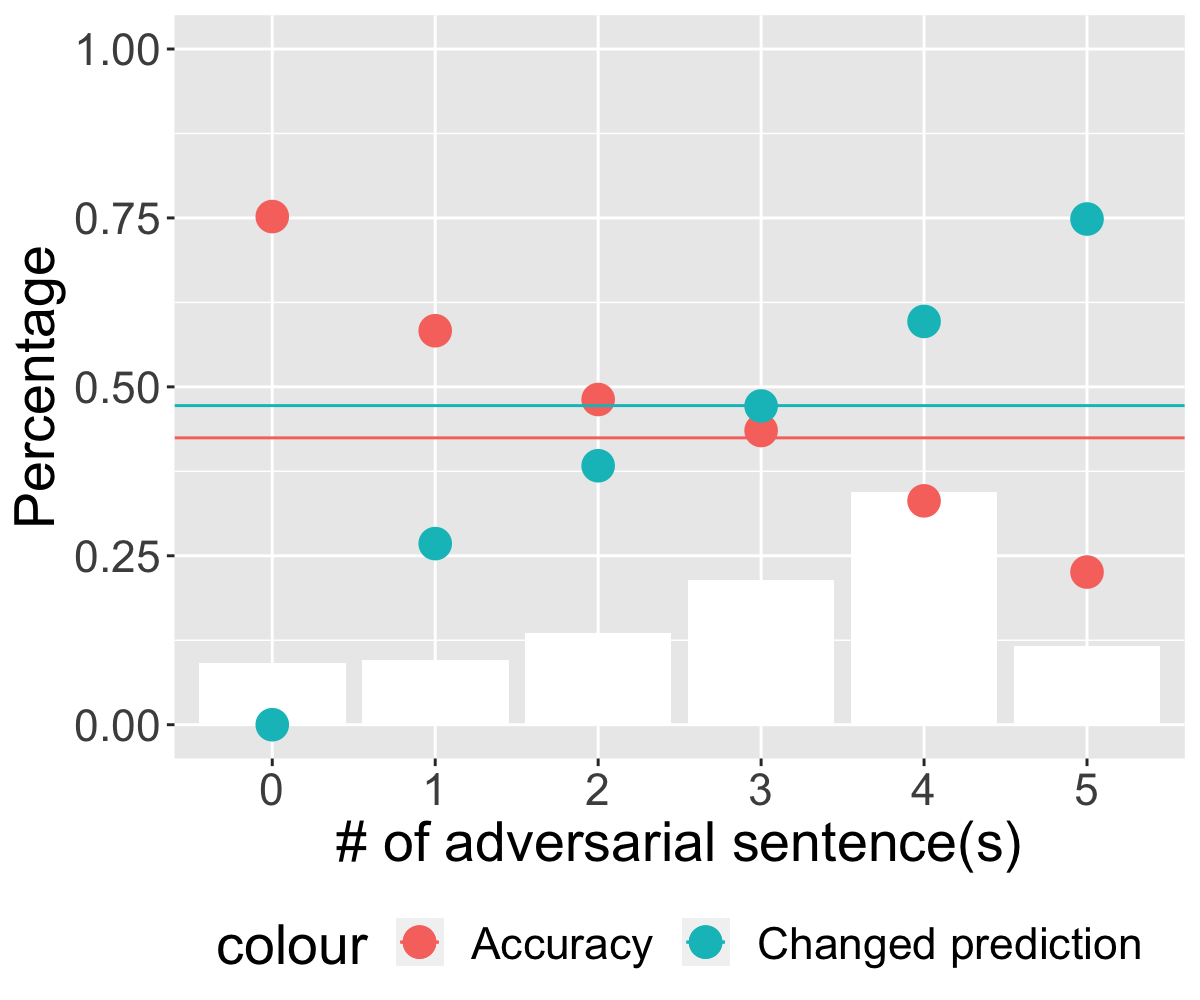}}
\hspace{\fill}
   \subfloat[\label{fig:sent-figure:nei_grover}NEI]{%
     \includegraphics[trim=10 0 0 0, clip,width=0.49\linewidth]{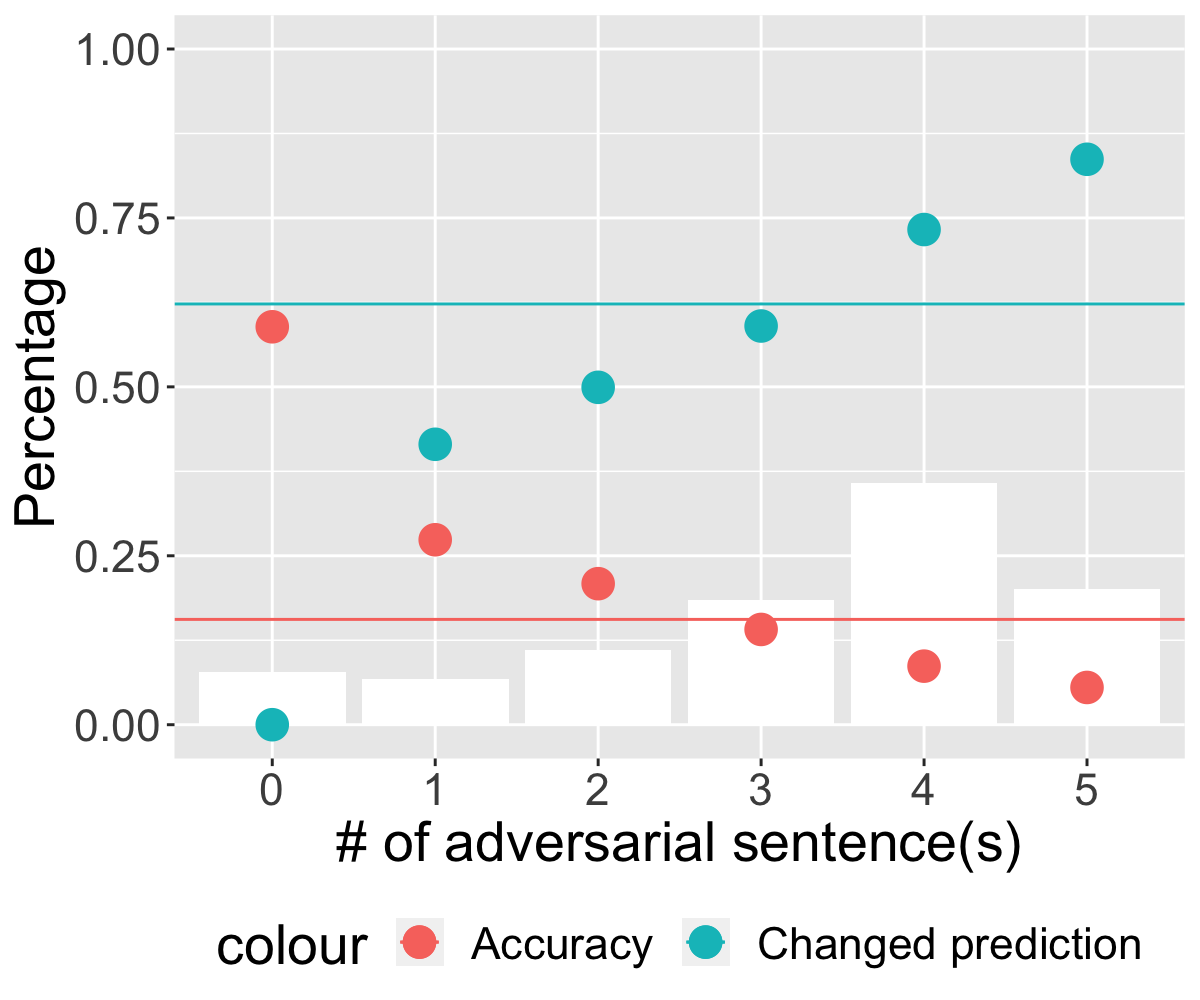}} \\
     \textbf{MediaCloud Evidence} \\
   \subfloat[\label{fig:sent-figure:ref_web}REF]{%
     \includegraphics[trim=10 0 0 0, clip,width=0.49\linewidth]{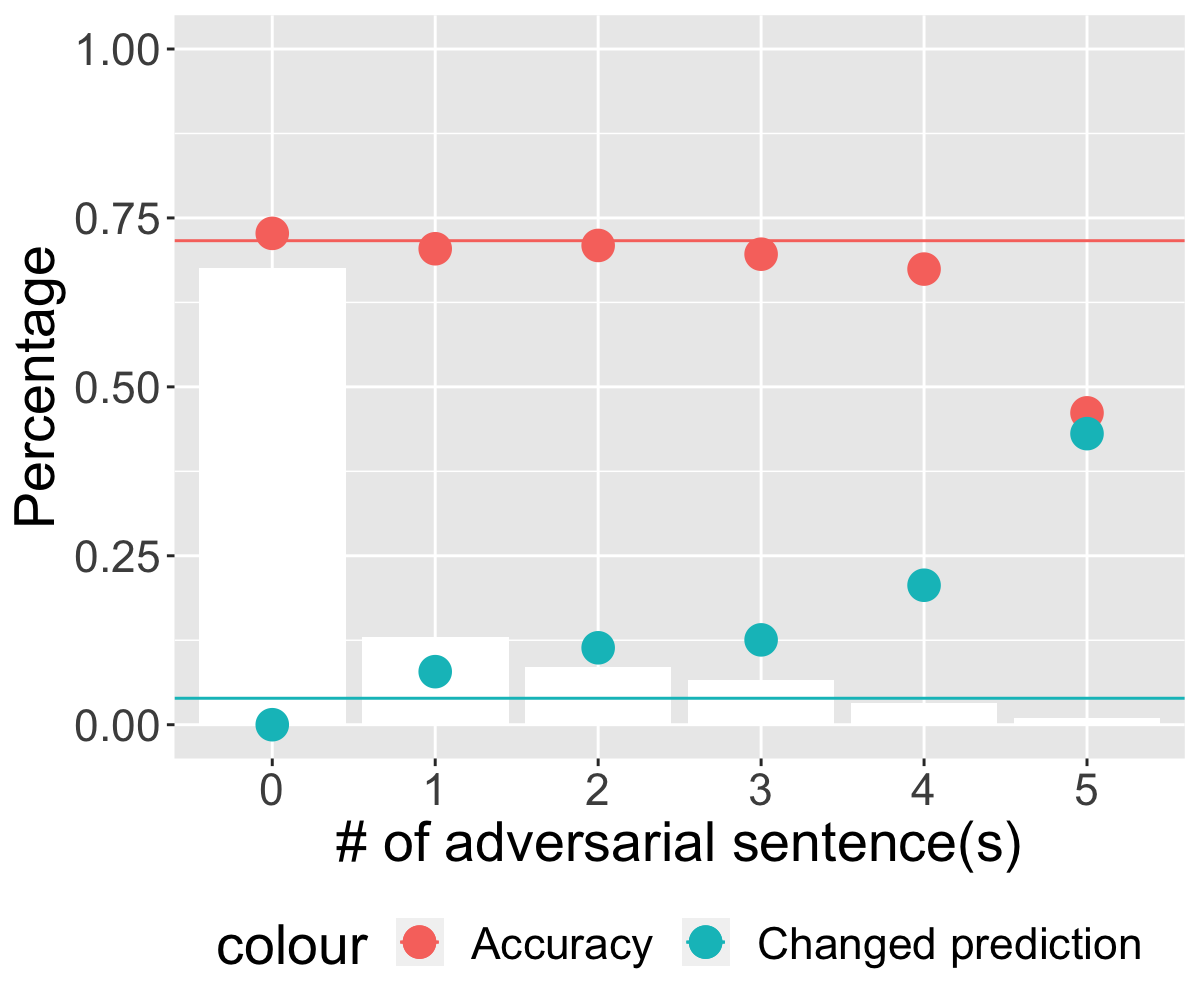}}
\hspace{\fill}
   \subfloat[\label{fig:sent-figure:nei_web}NEI]{%
     \includegraphics[trim=10 0 0 0, clip,width=0.49\linewidth]{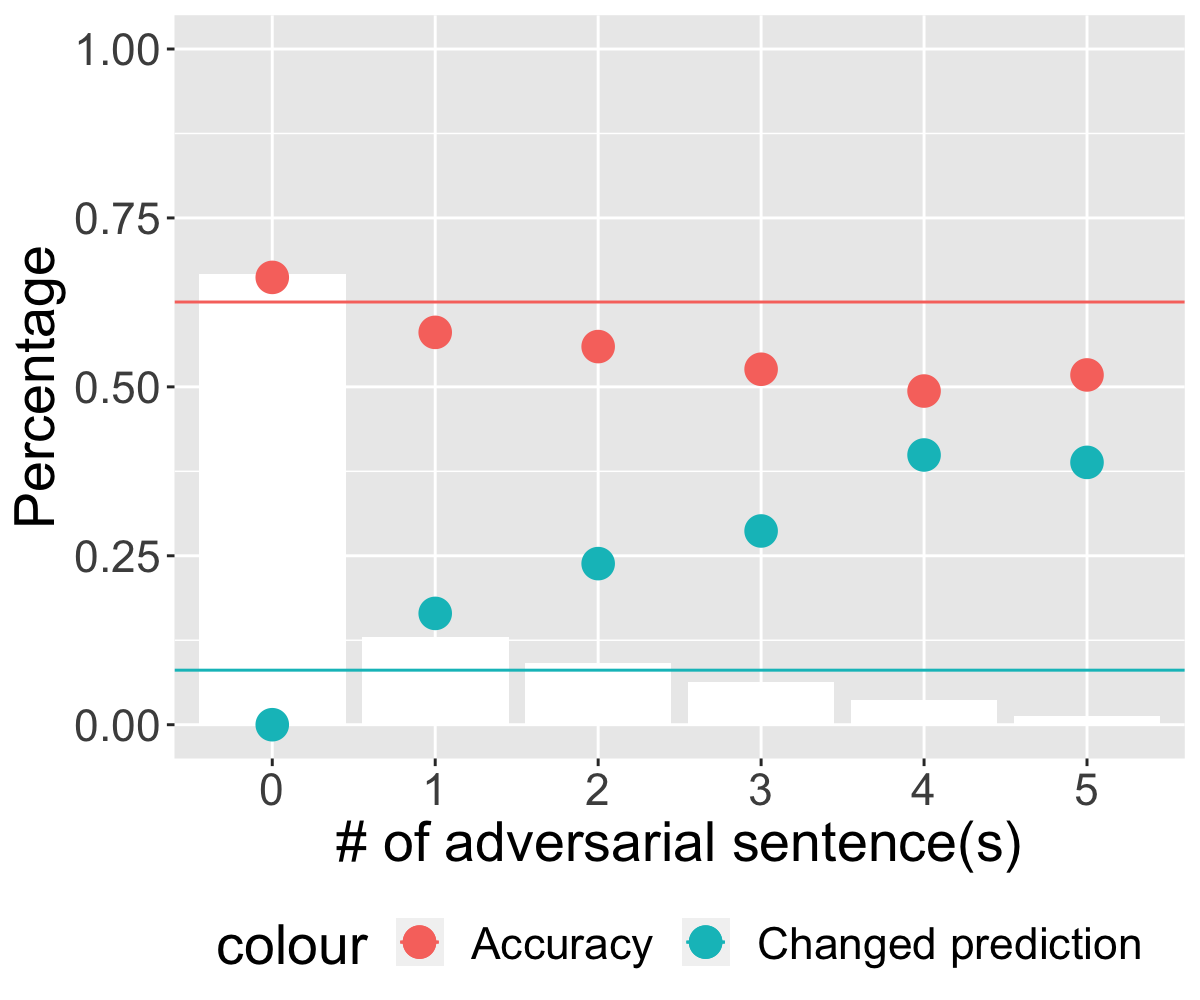}}
  \centering
  \caption{Degree of evidence poisoning and resulting REF (a,c) and NEI (b,d) claim verification accuracy for \advadd-\grover{}  (a,b) \advadd-MediaCloud (c,d)}
  \label{fig:sent-figure}
\end{figure}

Moreover, Figure~\ref{fig:sent-figure} shows a histogram of the number of poisoned evidence sentences retrieved per claim and a stratified analysis of the final predictions. Figures \ref{fig:sent-figure} (a)(b) for \advadd-\grover{} (\ie, \advadd-\textit{full}) show that accuracy steadily drops for claims with more corresponding evidence from poisoned documents; meanwhile, the likelihood that a prediction changes  increases with more poisoned evidence. We note that claims labeled NEI are far more sensitive to the introduction of poisoned sentences than REF claims, even as the rate of contamination is approximately the same between both types of labels. While this result is promising because the model is more robust in the presence of even minimal \textit{refuting} evidence, it also demonstrates that fact verification systems are more sensitive 
when no competing information is presented to a particular viewpoint 
(\ie, data voids; \citealp{Boyd2018DataVW}).     

\paragraph{Quality of poisoned evidence} We also evaluate whether poisoned evidence produced by \advadd{} is of sufficient quality to bypass potential human detectors. For 500 REF and 500 NEI claims from \fever, we ran a study on Mechanical Turk where we presented three workers with five retrieved evidence examples (which could be from \advadd{} or from \fever DB) and asked them to identify which examples were machine-generated. Our results show that humans underestimate the number of poisoned sentences (23.6\% recall), and do not distinguish machine- from human-generated evidence (48.6\% precision). While well-trained workers will improve at recognizing synthetic content, our results demonstrate the challenge of distinguishing these evidence sources for lay readers, pointing to the quality of the synthetic content, and the potential for such an attack to remain undetected.

\begin{table}[t]
    \centering
    \begin{tabular}{lrr}
        \toprule
        \multirow{2}{*}{\textbf{Source}}  &        \multicolumn{2}{c}{\textbf{Evidence Retrieval}} \\
         & \textbf{Document} & \textbf{Sentence} \\
        \midrule
        \grover{} & 87\% & 65\% \\
        MediaCloud & 99\% & 17\% \\
        \bottomrule
        && \\
        \toprule
        \multirow{2}{*}{\textbf{Source}} &        \multicolumn{2}{c}{\textbf{Claim Verification}} \\
         & \textbf{KGAT} & \textbf{CorefBERT} \\
        \midrule
        Original & 70.76 & 73.05 \\
        \midrule
        + \grover{} & 29.02 & 28.59 \\
        + MediaCloud & 67.10 & 70.44 \\
        \bottomrule
    \end{tabular}
    \caption{Statistics and performance relative to the source of poisoned evidence: \grover{} or MediaCloud}
    \label{tab:chosen-table}
    
\end{table}

\begin{table*}[t]
    \centering
    \begin{tabular}{llrrrr}
    \toprule
        \multirow{2}{*}{\textbf{Model}} & \multirow{2}{*}{\textbf{Evidence Set}} & \textbf{Sentence} & \textbf{Sentence} & \textbf{Abstract} & \textbf{Abstract} \\
        & & \textbf{selection} & \textbf{label} & \textbf{label} & \textbf{rationalized} \\
        \midrule
        VeriSci  & Original  & 47.69 & 42.62 & 51.03 & 48.45 \\
               \citep{wadden-etal-2020-fact} & \advadd  & \textbf{27.05} & \textbf{23.50} & \textbf{25.57} & \textbf{24.33} \\
               \midrule
        SciKGAT & Original & 55.61 & 51.69 & 58.04 & 57.41 \\
        \citep{liu-etal-2020-adapting-open} & \advadd &  \textbf{39.44} & \textbf{36.97} & \textbf{37.46} & \textbf{36.98} \\
                \midrule
        ParagraphJoint  & Original  & 53.63 & 43.59 & 55.52 & 49.55 \\
        \citep{li2021paragraphlevel} & \advadd  & \textbf{37.68} & \textbf{32.60} & \textbf{41.31} & \textbf{37.12} \\
    \bottomrule
    \end{tabular}
    \caption{Effect of \advadd{} evidence on the \scifact{} benchmark. We \textbf{bold} performance drops relative to the original evidence. 
    }
    \label{tab:scifact}
\end{table*}

\paragraph{Comparison with human-compiled online evidence}
While we have shown that synthetic disinformation affects the performance of downstream claim verifiers when present in their evidence sets, the threat should be evaluated in comparison to the threat of already existing online misinformation on the same topic.
Consequently, we use the MediaCloud\footnote{mediacloud.org} content analysis tool to crawl web content related to \fever{} claims. We crawl English-language news since January 2018 that contains the keywords of a claim anywhere in their text and extract articles with a title that contains at least one keyword from the claim. Finally, we process these articles to have the same format as the original Wikipedia database, yielding 74M total available documents for retrieval (Table~\ref{tab:repo-stats}). 

In Table~\ref{tab:chosen-table}, we report the performance of an \advadd{} setting where only MediaCloud-crawled documents are available to the retriever compared to our original setting where \grover-generated documents were available. 
We observe that evidence crawled from online content has less of an influence on downstream fact verification performance ($\sim$3\% vs. $\sim$40\% performance drop). While we are able to retrieve far more documents from MediaCloud due to the size of the database (99\% of claims retrieve a document from MediaCloud), the sentences from \advadd-\grover{} documents are selected more frequently in the sentence retrieval step. While this gap would likely be less pronounced with more contentious claims that yield competing viewpoints \citep{bush_zaheer_2019}, these results demonstrate that synthetic disinformation can be much more targeted to a particular claim of interest. Figure \ref{fig:sent-figure} supports this conclusion, where we observe smaller performance drops for \advadd-MediaCloud (c,d) compared to \advadd-\grover{} (a,b) even when all retrieved sentences are sourced from the poisoned evidence.

\subsection{\scifact{} Study}

\paragraph{Setup}
For \scifact{}, we chose three systems for testing our attack: VeriSci \citep{wadden-etal-2020-fact}, ParagraphJoint \citep{li2021paragraphlevel}, and SciKGAT \citep{liu-etal-2020-adapting-open}. The VeriSci model was released by the creators of the \scifact{} benchmark and retrieves relevant abstracts to a claim using TF-IDF. The ParagraphJoint model, one of the top systems on the \scifact{} leaderboard, uses a word embedding-based method to retrieve abstracts. Both use a RoBERTa-based module for rationale selection and label prediction. 
The SciKGAT model uses the same evidence retrieval as VeriSci, but the KGAT model \citep{liu2020kernel} to verify claims.
We use the 300 claims from the development set to evaluate our method.  
We generate \grover{} articles as with \fever, but we set the venue URL to \url{medicalnewstoday.com}, which produces articles more likely to reflect scientific and medical content. 

\paragraph{Results} In Table~\ref{tab:scifact}, we observe large performance drops across all metrics for all models. Furthermore, we note that our disinformation generator, \grover, is not trained on large quantities of scientific documents of the same format as the original evidence. Despite producing documents that are stylistically different, the disinformation is still retrieved as evidence, and affects the performance of the verifier.

\subsection{\covidfact{} Study}

\paragraph{Setup} We run our analysis on the baseline system from \citet{Saakyan2021COVIDFactFE}. This model retrieves evidence documents for claims using Google Search and then selects evidence sentences based off high cosine similarity between sentence embeddings of the claims and individual evidence sentences \cite{reimers-gurevych-2019-sentence}. A RoBERTa-based model predicts a veracity label. We generate \advadd{} articles in the same manner as for \scifact, and run our analysis on the 271 REF claims from the test set.

\paragraph{Results} In both the Top-1 and Top-5 settings from \citet{Saakyan2021COVIDFactFE}, we observe a $\sim$14.4\% performance drop on REF claims (83.8\% $\rightarrow$  69.4\%). We note that \covidfact{} random and majority accuracy is only 67.6\% due to a label imbalance.

%% file: 5-sentence.tex
\begin{table*}[t]
  \centering
  \begin{tabular}{lrrrrrrrrrrrr}
    \toprule
    \multirow{3}{*}{\textbf{Evidence}}  &  \multicolumn{3}{c}{\textbf{CorefBERT Acc.}} & \multicolumn{3}{c}{\textbf{KGAT Acc.}} & \multicolumn{3}{c}{\textbf{MLA Acc.}} \\
    &  \multicolumn{3}{c}{\citep{ye2020corefbert}} & \multicolumn{3}{c}{\citep{liu2020kernel}} & \multicolumn{3}{c}{(Kruengkrai et al. 2021)}\\
         & \textbf{Total}  & \textbf{REF} & \textbf{NEI} & \textbf{Total} &  \textbf{REF} & \textbf{NEI} & \textbf{Total} & \textbf{REF} & \textbf{NEI}\\
    \midrule
    Original & 73.05 & 74.03 & 72.07 & 70.76 & 72.50 & 69.01 & 75.92 & 78.71  & 73.13 \\
    \midrule
    \advmodkey & 53.83 & 66.50 & 41.15  & 42.82 & 68.90 & \textbf{16.74} & 60.93 & 81.83  & 40.02  \\
    \advmodpara & \textbf{32.62} & \textbf{36.66} & \textbf{28.58} & \textbf{37.22} & \textbf{51.74} & 22.70 & \textbf{52.72} & \textbf{70.57} & \textbf{34.86} \\
    \midrule
    \advmod-\textit{oracle} (\textit{claim}) & 4.78 & 7.94 & 1.61  & 11.90 & 23.78 & 0.02 & 25.17 & 45.96 & 4.37 \\
    \bottomrule
  \end{tabular}
  \caption{Effect of \advmod{} on \fever{} claim verification. We \textbf{bold} the largest performance drop relative to the original evidence.} 
    \label{tab:advmod}
\end{table*}
  
\begin{table}[ht]
\resizebox{\linewidth}{!}{
    \centering
    \begin{tabular}{cc}
    \toprule
        \textbf{Original} & Damon Albarn's debut album was released in 2011. \\
        \midrule
        \multirow{2}{*}{\textbf{Paraphrase}} & \textbf{Albarn's first} album was released in 2011.\\
        & \textbf{His} debut album was released in 2011.\\
        \midrule
        \multirow{2}{*}{\textbf{KeyReplace}} & \textbf{Matt Coldplay}'s debut album was released in \textbf{202}. \\
        & \textbf{Stefan Blur}'s debut album was released in \textbf{1822}. \\
    \bottomrule
    \end{tabular}
    
}
    \caption{Sample \advmod{} sentences}
    \label{tab:advmod-example}
\end{table}

\section{\advmodlong: \\ Evidence Document Poisoning}
\label{sec:advmod}

In Section~\ref{sec:advadd}, we investigated the effect of adding disinformation documents to the evidence repositories of fact verification systems, simulating the setting where the dynamic pace of news might lead to fake information being used to verify real-time claims. However, even in settings where information has more time to settle and facts to crystallize, misinformation can still find its way into repositories of documents used by fact-checking systems. 
Motivated by the possibility of malicious edits being made to crowdsourced information resources, we explore how NLP methods could be used to automatically edit existing articles with fake content at scale.

\subsection{Approach}

Our method, \advmodlong{} (\advmod), simulates this setting in a two-stage process. First, we use off-the-shelf NLP tools to generate modified versions of the claim presented to the fact verifier. Then, we append our modified claims to articles in the evidence base that are relevant to the original claim. We modify the original claims in two ways.

In the \textit{paraphrase} approach, we use a state-of-the-art paraphrasing model, PEGASUS \citep{zhang2019pegasus}, to generate paraphrased versions of the original claim (see Table~\ref{tab:advmod-example} for examplea). These paraphrases generally retain the meaning of the claim, but often remove contextualizing information that would be found in the context of the article in which the new sentence is inserted. Because the \textit{paraphrase} method attempts to produce synthetic evidence that is semantically equivalent to the original claim, we test its efficacy relative to a method that merely introduces irrelevant content to the evidence document. Motivated by \citet{DBLP:conf/emnlp/JiaL17}, we alter a claim by applying heuristics such as number alteration, antonym substitution, and entity replacement with close neighbors according to embedding similarity \citep{bojanowski2017enriching}. 
These modifications should not confuse humans, but would affect sensitive fact verification systems, providing a competitive baseline for assessing the strength of \advmod-\textit{paraphrase}. 

Finally, our oracle reports the performance when the claim itself is appended to an evidence document.

\subsection{Results}

Our results in Table~\ref{tab:advmod} demonstrate that injecting poisoned evidence sentences into existing documents is an effective method for fooling fact verification systems. Our \advmodpara{} method causes a significant drop on all tested models for both REF and NEI labeled claims. Furthermore, we also note that \advmodpara{} achieves larger performance drops than the baseline method, \advmodkey, for most claim types (the KGAT model is slightly more sensitive to the baseline \advmodkey{} for the NEI claims), indicating that injections of disinformative content are more effective than non-targeted perturbances to the evidence (\eg, \advmodkey). 

%% file: 6-discussion.tex
\section{Discussion}
\label{sec:discussion}

Adding synthetic content to the evidence bases of fact verifiers significantly decreases their performance. Below, we discuss interesting findings and limitations of our study.

\paragraph{Synthetic vs. human disinformation}

As mentioned in Section~\ref{sec:advadd}, the performance of our test systems is more sensitive to poisoned evidence generated from \grover{} than retrieved from MediaCloud, even as the number of documents retrieved from MediaCloud far exceeds the number generated from \grover. While \fever{} claims may not generally be worth opposing online (leading to less directly adversarial content being retrieved from MediaCloud), we note that language models have no such limitations, and can generate large quantities of disinformation about any topic. Consequently, while misinformation already makes its way into retrieval search results \citep{bush_zaheer_2019}, language models could cheaply skew the distribution of content more drastically \citep{Bommasani2021OnTO}, particularly on topics that receive less mainstream coverage, but may be of import to a malicious actor \citep{Starbird2018}.

\paragraph{Language models as a defense} In the \advadd{} and \advmod{} oracle settings, all tested systems performed better on claims labeled REF than for claims labeled NEI. This result implies that the \grover-generated evidence was less adversarial for these claims, or that the pretrained models which these systems use to encode the claim and evidence sentences were more robust against claims that should be \textit{refuted}. Consequently, we conclude that language models encode priors about the veracity of claims, likely from the knowledge they learn about entities during pretraining \citep{Petroni2019LanguageMA}, a conclusion also supported by contemporaneous work in using standalone language models as fact-checkers \citep{Lee2020LanguageMA,Lee2021TowardsFF}. While this property can be an advantage in some settings (\ie, language models pretrained on reliable text repositories will be natural defenses against textual misinformation), it will also be a liability when previously learned erroneous knowledge will counteract input evidence that contradicts it. Finally, we note that the presence of implicit knowledge in language models affecting the interpretation of input evidence implies that the training corpora of these LMs could be attacked to influence downstream fact verification. Prior work has explored poisoning task-specific training datasets \citep{wallace2021concealed}. As disinformation becomes more prevalent online, the pretraining corpora of LMs will require more careful curation to avoid learning adversarial content.

\paragraph{Limitations} We identify three main limitations to our study. First, the \fever{} document retrievers use the MediaWiki API to collect relevant Wikipedia articles based on entity mentions in the claim. We assume our synthetic content could be included in the retrieved documents if it were titled with a mention of the named entities in the claim. For \scifact{}, this limitation is not present because synthetic abstracts are retrieved using statistical IR methods. Second, our method \advadd{} uses the actual claim to generate the synthetic article. In the absence of explicit coordination, synthetic poisoned evidence would be generated without knowledge of the exact claim formulation, reducing the realistic correspondence between the claim and the synthetic disinformation. 
If the \grover{} model directly copied the claim during generation, performance drops might be overestimated based on unrealistically aligned evidence. 
For the \advadd{}-\textit{full} setting, we measure that this issue arises in $\sim$20\% of claims, which are predicted incorrectly more often, but does not affect the conclusions of our study. Finally, our \fever{} and \covidfact{} studies are run using only claims labeled as REF and NEI, which we discuss in more detail in the Appendix.

%% file: 7-conclusion.tex
\section{Conclusion}

In this work, we evaluate the robustness of fact verification models when we poison the evidence documents they use to verify claims. We develop two poisoning strategies motivated by real world capabilities: \advadd, where synthetic documents are added to the evidence set, and \advmod, where synthetic sentences are added to existing documents in the evidence set. Our results show that these strategies significantly decrease claim verification accuracy. 
While these results are troubling, we are optimistic that improvements in automated synthetic content detection, particularly by online platforms with considerable resources, combined with human audits of fact-checker evidence (and their source), may still potentially mitigate many attempted disinformation campaigns.

%% file: 8-appendix.tex
\appendix

\section{Additional \advadd{} Results} 
\label{sec:app:advadd}
\paragraph{Effect of increased evidence:} When we increase the number of evidence sentences from 5 to 10 during the claim verification step, we see minimal difference in the performance drop. When only one adversarial sentence is inserted the performance drop decreases from 23.87\% for 5 sentences to 23.69\% for 10 sentences for REF claims. For NEI claims, the performance drop actually increases from 49.49\% (5 sentences) to 50.81\% (10 sentences).

\paragraph{Effect of random evidence:} The performance drops reported in Section~\ref{sec:advadd} might be due to correct evidence being removed from the retrieved set, rather than poisoned evidence being introduced. To test this possibility, we ran an experiment where we replace the retrieved poisoned evidence sentences with randomly chosen sentences from \fever DB. If poisoned evidence does not adversely affect the claim verifier beyond the replacement of potentially useful supporting sentences, we should expect minimal performance drop from this baseline. However, we find that when random sentences are inserted, the performance drop for the KGAT model shrinks from 30.05\% to 6.63\% for REF claims and from 53.42\% to 7.73\% for NEI claims. Similarly, in a proxy for the \advadd-\textit{min} setting, where only a single sentence is replaced, the shrink is from 28.87\% to 1.86\% for REF claims and from 49.49\% to 7.01\% for NEI claims. These results demonstrate that the performance drop comes from the addition of adversarial evidence instead of only the removal of possibly correct evidence, indicating that the claim verifier is directly sensitive to the content of poisoned evidence. 

\begin{table}[h!]
    \centering
    \small
    \begin{tabular}{l|rrrr|r}
        \toprule
        & \multicolumn{4}{c|}{\textbf{Claim Verifier}} & \textbf{\%}\\
        \textbf{Sentence}  &     \multicolumn{2}{c}{\textbf{REF Claims}} &  \multicolumn{2}{c|}{\textbf{NEI Claims}} & \textbf{Corrupt}  \\
        \textbf{Retriever} $\downarrow$ & \textbf{MLA} & \textbf{KGAT} & \textbf{MLA} & \textbf{KGAT} & \textbf{Sents}\\
        \midrule
        \textbf{MLA} & 71.84 & \textbf{\color{blue}{49.89}} & 31.87 & \textbf{\color{blue}{18.18}} & 50.1 \\
        \textbf{KGAT} & \textbf{\color{red}{23.42}} & 42.45 & 30.24 & 15.59 & 62.5 \\
        \bottomrule
    \end{tabular}
    \caption{Effect of the sentence retriever on MLA and KGAT.}
    \label{tab:retriever-swap}
    
\end{table}

\begin{figure}[t!]
    \centering
    \textbf{KGAT} \\
   \subfloat[\label{fig:advmod-additional-figure:keyreplace_kgat}KeyReplace]{
     \includegraphics[trim=10 0 0 0, clip, width=0.48\linewidth]{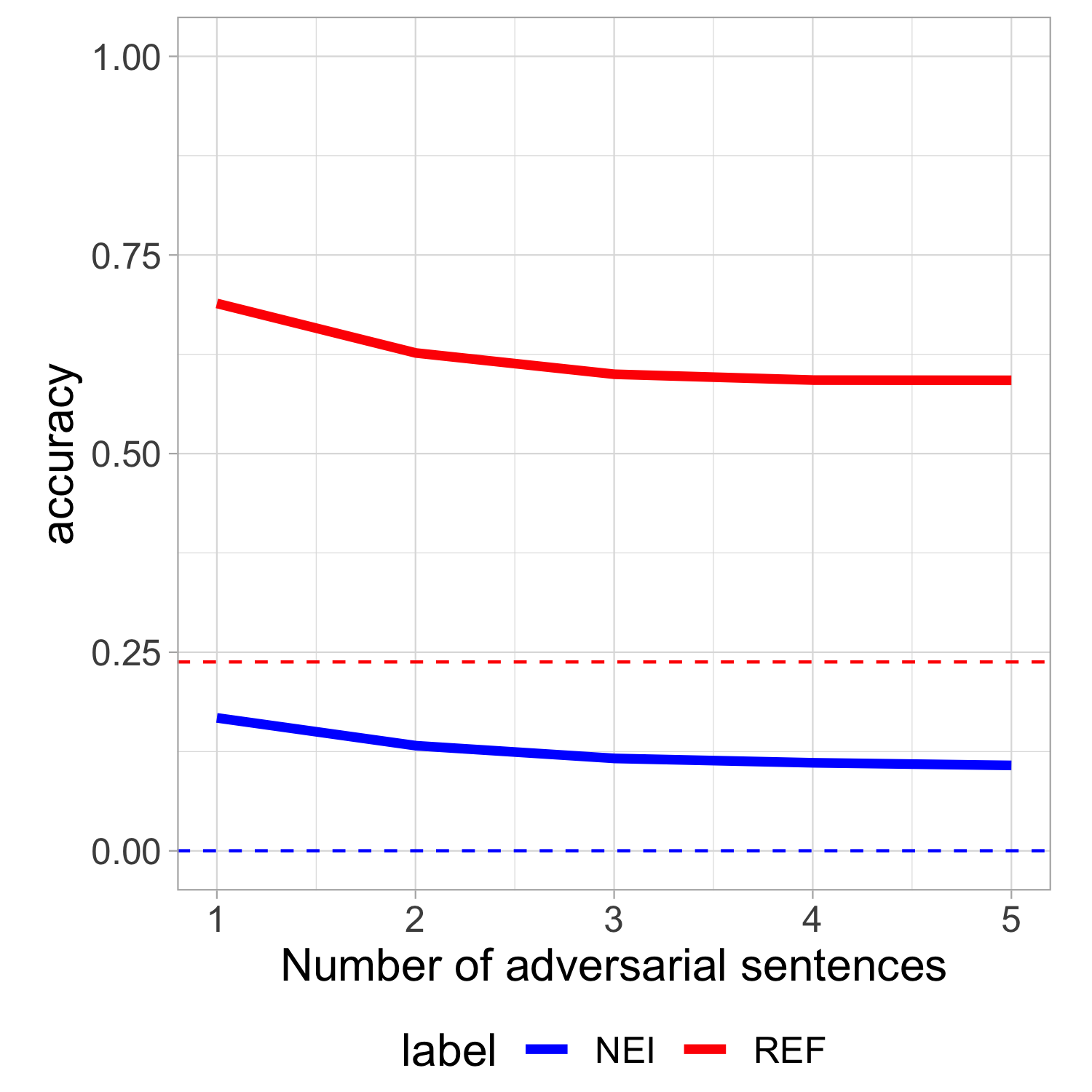}}
\hspace{\fill}
   \subfloat[\label{fig:advmod-additional-figure:paraphrase_kgat}Paraphrase]{
     \includegraphics[trim=10 0 0 0, clip,width=0.48\linewidth]{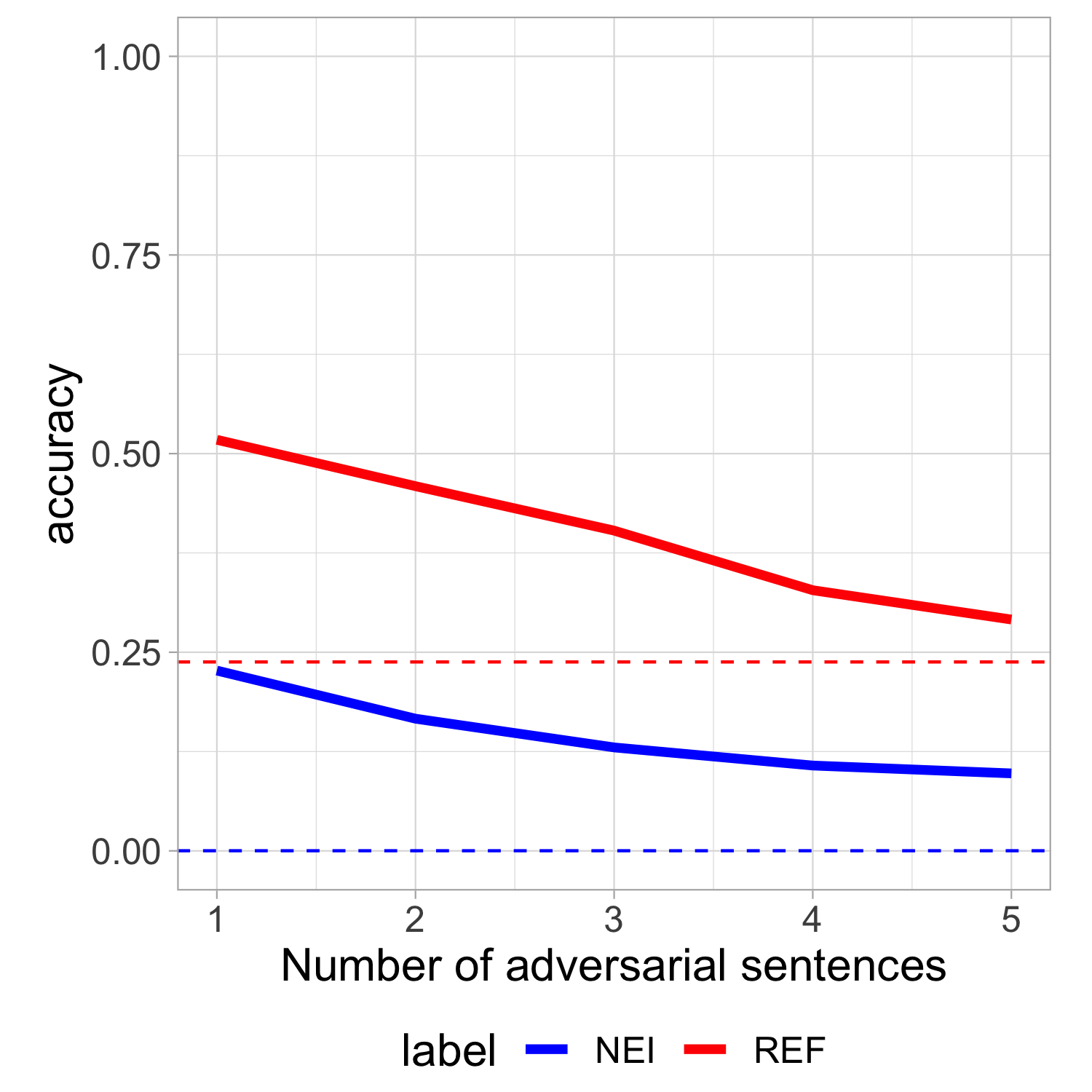}} \\
     
    \textbf{CorefBERT} \\
   \subfloat[\label{fig:advmod-additional-figure:keyreplace_corefbert}KeyReplace]{
     \includegraphics[trim=10 0 0 0, clip, width=0.48\linewidth]{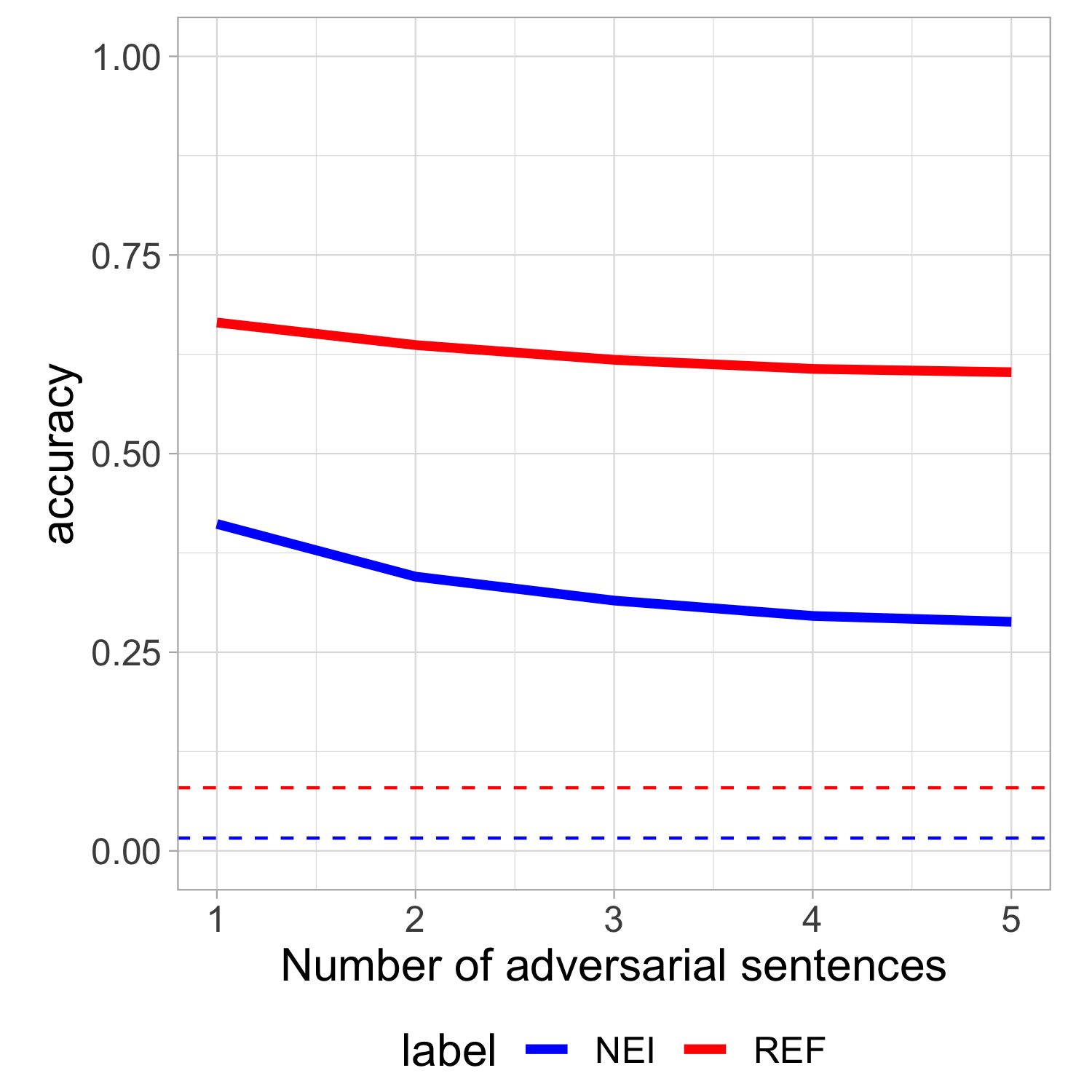}}
\hspace{\fill}
   \subfloat[\label{fig:advmod-additional-figure:paraphrase_corefbert}Paraphrase]{
     \includegraphics[trim=10 0 0 0, clip,width=0.48\linewidth]{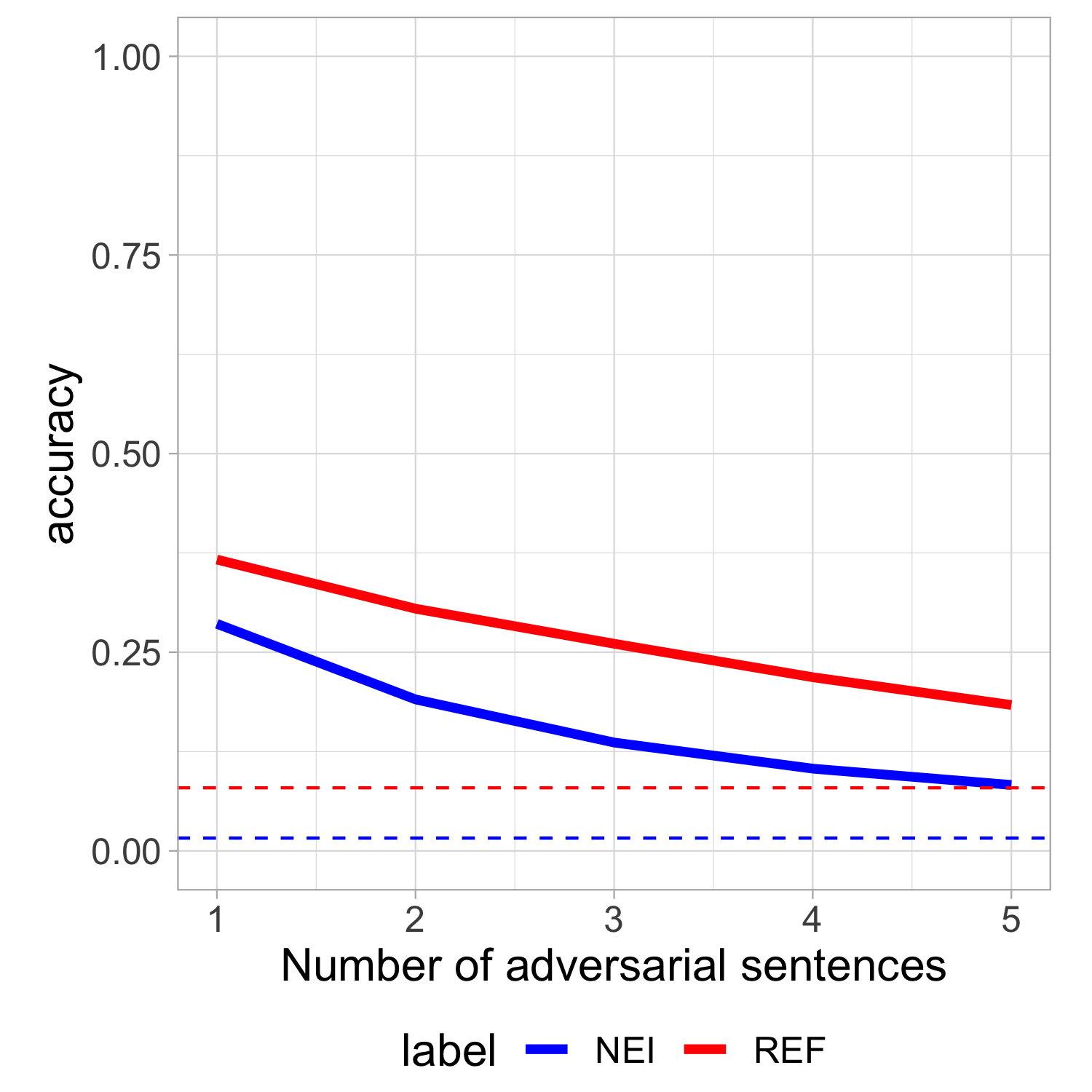}} \\
     
     \textbf{MLA} \\
   \subfloat[\label{fig:advmod-additional-figure:keyreplace_mla}KeyReplace]{
     \includegraphics[trim=10 0 0 0, clip, width=0.48\linewidth]{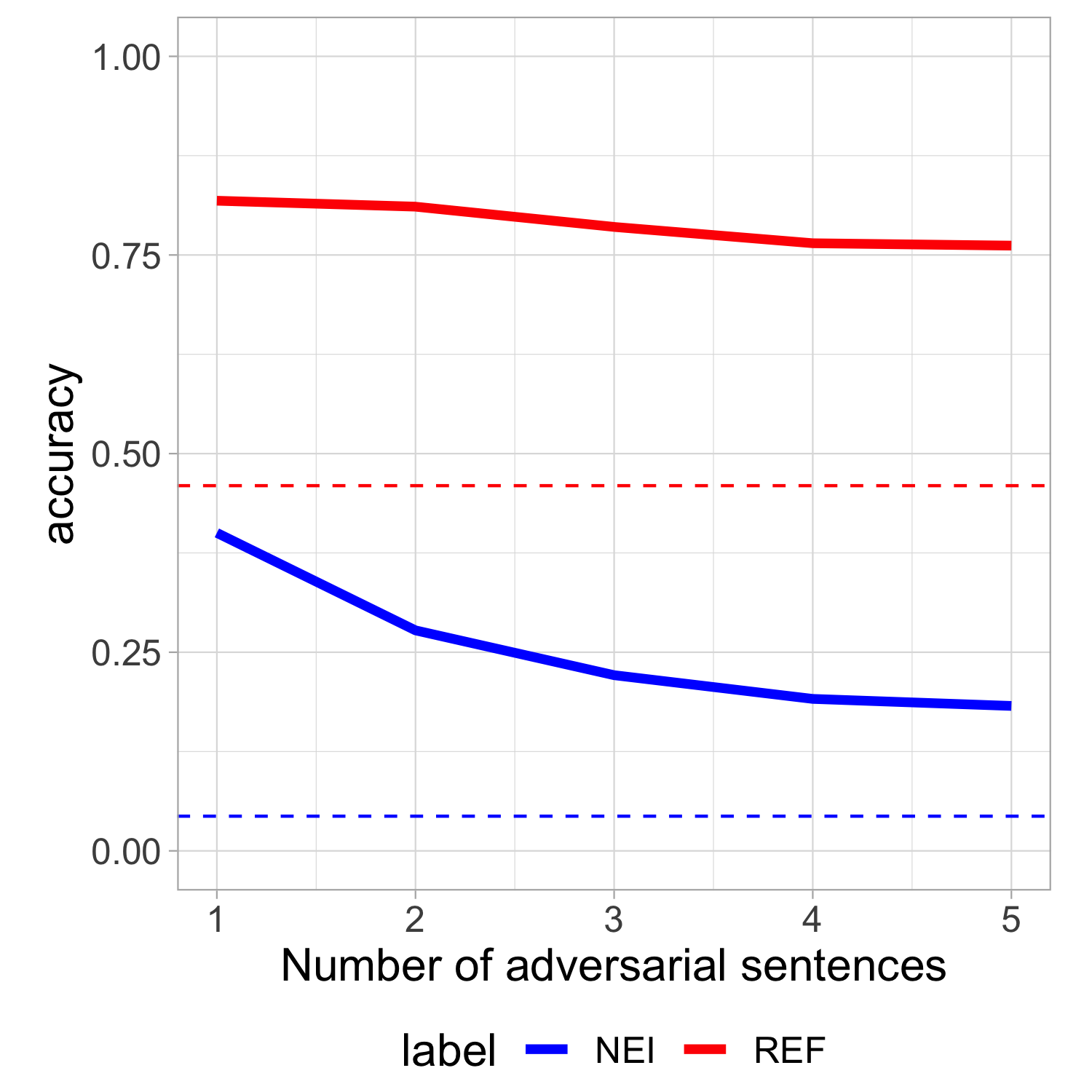}}
\hspace{\fill}
   \subfloat[\label{fig:advmod-additional-figure:paraphrase_mla}Paraphrase]{
     \includegraphics[trim=10 0 0 0, clip,width=0.48\linewidth]{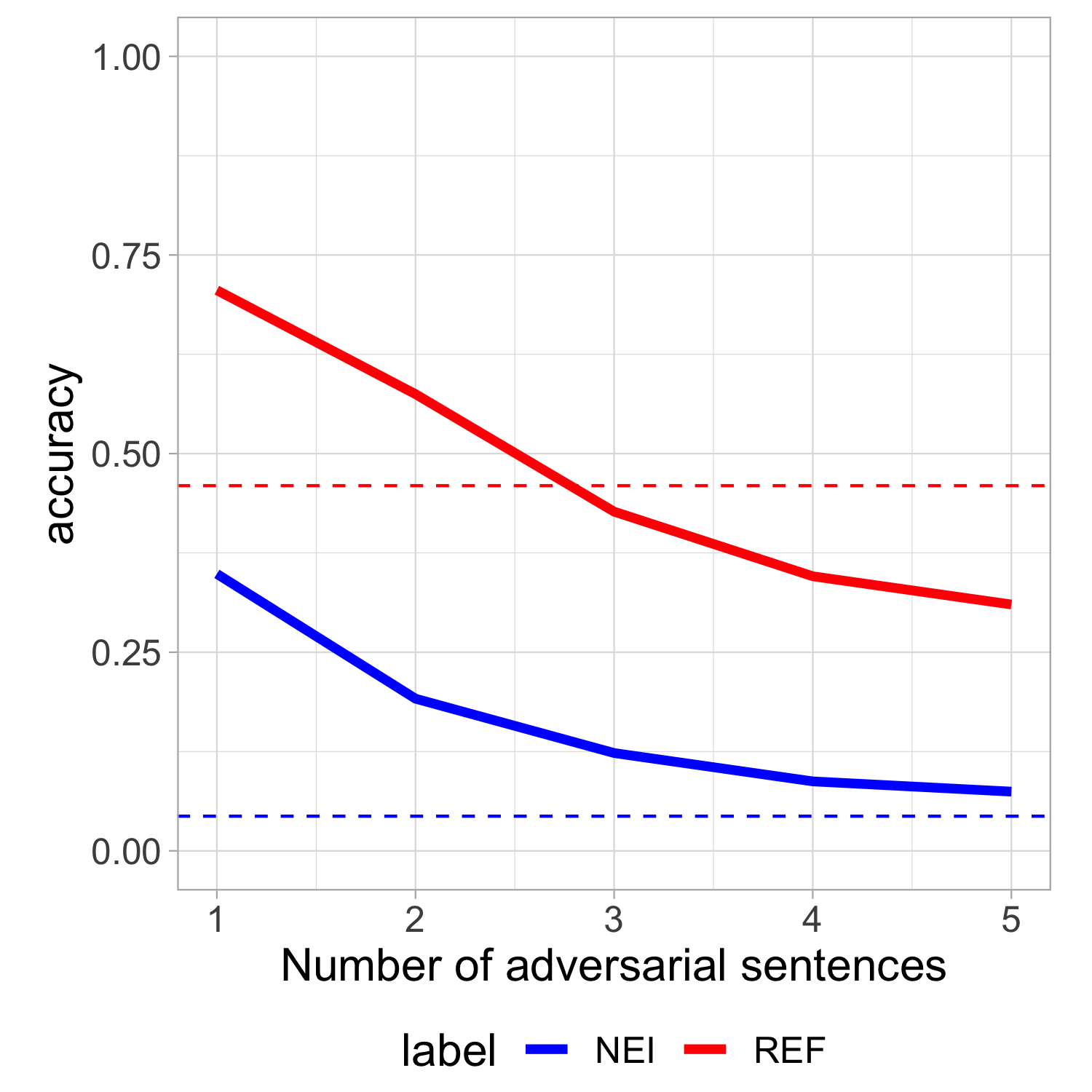}} \\
  \centering
  \caption{Claim verification accuracy by model and poisoning technique for different amounts of \advmod{} evidence poisoning. The dashed line shows \advmod{} performance when the claim itself is added as a single evidence sentence.}
  \label{fig:advmod-additional-figure}
\end{figure}

\paragraph{Effect of sentence retrieval performance} Our results in Table~\ref{tab:advadd-perf} show the MLA model \citep{Kruengkrai2021AMA} is more robust to poisoned evidence. To explore the cause of this finding, we swap the sentence retrievers of the KGAT and MLA models to disentangle the contributions of their sentence retrievers and claim verifiers.  
In Table~\ref{tab:retriever-swap}, we find that when the MLA sentence retriever is paired with the KGAT claim verifier, the performance of this joint system (highlighted in \textbf{\color{blue}{blue}}) increases relative to using the full KGAT model. Meanwhile, when paired with the KGAT sentence retriever, the MLA claim verifier achieves lower performance on REF claims (highlighted in \textbf{\color{red}{red}}) than the full KGAT model, indicating that the strength of the MLA model may stem from a more powerful retriever. However, the MLA claim verifier is also more robust for NEI claims regardless of the retriever, implying that the claim verifiers of these models may suffer from \textit{exposure bias}, 
whereby they overfit to their sentence retrievers during training. They learn to expect certain patterns from the evidence returned by these retrievers and when given evidence from another distribution (\ie, a different retriever) at test time, they perform worse on examples that require retrieval (\ie REF claims).

\section{Additional \advmod{} Results}
In Figure~\ref{fig:advmod-additional-figure}, we report additional \advmod{} results measuring the performance change as a function of the amount of documents we re-write in the document base. Our results show that editing multiple documents is more likely to cause the prediction to flip. However, even a single edit to an evidence document can often cause a large performance drop.

\begin{table*}[t]
    \centering
    \begin{tabular}{rl}
    \toprule
         \textbf{Original SUP Claim} & Ron Weasley is part of the Harry Potter series as the eponymous wizard's best friend. \\
         \midrule
         \textbf{Manual Counterclaim} & Ron Weasley is part of the Harry Potter series as the eponymous wizard's worst enemy. \\
         \textbf{\grover{} Output} & According to the 'Harry Potter: The Story of Ron Weasley' campaign , the prince of Goblet \\
         & of Fire and Hogwarts castle is part of the series as the eponymous wizard's worst enemy .\\
         \midrule
         \textbf{Automatic Counterclaim} & Ron Weasley is part of the Harry bee series as the eponymous wizard's best friend. \\
         \textbf{\grover{} Output} & Unlike the remainder of Harry Potter , Ron Weasley is more than just a character on the \\
         & Hogwarts kiddie roster. \\
         \midrule
         \midrule
         \textbf{Original SUP Claim} & Bessie Smith was a singer. \\
        \midrule
        \textbf{Manual Counterclaim} & Bessie Smith cannot sing. \\
        \textbf{\grover{} Output} & Bessie Smith , a kindergarten teacher in Chicago , Illinois , is scheduled to perform for the \\
        & 2014 Eagles open game against the Giants on Monday Night Football (18:00 ET on ESPN) .\\
        \midrule
        \textbf{Automatic Counterclaim} & Bessie Smith was a vegetarian. \\
        \textbf{\grover{} Output} & So how did Bessie Smith become a vegetarian ? The black American woman sat on the \\
        & United States House Floor as a member of the Congressional Choir during the 70s , when , \\
        & she actually was a vegan. \\
         \bottomrule
    \end{tabular}
    \caption{Sample \advadd{} document excerpts generated by \grover{} for SUP claims in the \fever{} dataset.}
    \label{tab:sup-sample}
\end{table*}

\section{Performance on \textit{Supports} Labels}
Our studies on the \fever{} and \covidfact{} benchmarks focused on the claims labeled \textit{refutes} (REF) and \textit{not enough information} (NEI). Claims labeled as \textit{supports} (SUP) were not included in the study due to the challenge of generating effective poisoned evidence for them.  Generating poisoned evidence for \fever{} NEI and REF claims is more straightforward because we can use variants of the claim (\eg, paraphrases) or the claim itself as input to \grover{} to produce poisoned evidence. However, poisoned evidence can only be generated for SUP claims if suitable counterclaims can be formulated as an input to \grover{}.

To test our method on SUP claims, counterclaims were generated in the following manner: we adapted the automatic counterclaim generation method from \citet{Saakyan2021COVIDFactFE}, which selects salient words in the original claim using an aggregate attention score for each token based on how much it is attended to by the other tokens in the sequence. Then, the most salient token is replaced by sampling from a masked language model. Once we generate a set of counterclaims using this method, we validate them using the decomposable attention NLI model of \citet{parikh-etal-2016-decomposable} by selecting the ones with the highest contradiction score with respect to the original claim. Then, we provided these counterclaims as inputs to \grover{} to generate poisoned evidence. However, we found this method was not effective for generating poisoned evidence. When we ran our \advadd{} setting using the KGAT model, we observed a surprising performance increase from 86.2\% to 87.5\% on label prediction accuracy, indicating that the generated poisoned evidence unexpectedly helps the model make correct predictions. 

Examples in Table~\ref{tab:sup-sample} depict the limitations of this approach. In the first example, the change made to generate the counterclaim does not change the semantics of the claim, merely changing the word ``Potter'' to ``bee,'' which is not a coherent coherent counterclaim that would produce poisoned evidence from \grover{} refuting the original claim. In the second example, the counterclaim is coherent, but does not semantically contradict from the original claim, making the poisoned evidence less likely to be retrieved when the original claim is provided to the fact verification model. Furthermore, we note the difficulty of generating counterclaims for many claims in \fever{}. First, many of the original claims are not easily falsifiable (\eg, ``Girl is an album''), making it challenging to formulate a suitable counterclaim. Other are statements that cannot be falsified without using explicit negation terms (\eg, ``Stripes had a person appear in it''). As language models struggle to understand inferences of negated statements \citep{kassner-schutze-2020-negatedprobes,jiang-etal-2021-im}, \grover{} may just as often generate content that ends up supporting the original claim, rather than contradicting it, when seeded with such explicitly negated counterclaims.

However, the focus of our study is whether synthetically-generated adversarial evidence could be generated at scale to mislead fact verification systems. While generating counterclaims automatically at scale is necessary to perform this study on \fever{} SUP claims, an adversary would be more likely to generate synthetic content for a single claim (or related claims) of interest (rather than a large set). Consequently, they would be able to manually write the counterclaim that was needed to generate poisoned evidence, mitigating the need for automatic counterclaim generation methods. We evaluate this possibility by writing counterclaims for a sample of 100 \fever{} SUP claims, allowing us to guarantee semantic contradiction of the original claim by the counterclaim (as seen in Table~\ref{tab:sup-sample}). However, we find that, once again, performance does not drop as the original performance on these claims was 92\% and rose to 93\% once the poisoned evidence from \grover{} was available. Though manually writing contradicting statements guarantees coherence and quality of the counterclaim, \grover{} may still fail to generate content as intended and may even affirm the original claim, possibly because the model has been trained on Wikipedia, indicating that \grover{} may encode implicit knowledge about many of the entities for which it must produce poisoned evidence, as discussed in Section~\ref{sec:discussion}.  
For example, we note that one of the counterclaims from Table~\ref{tab:sup-sample} --- ``Bessie Smith was a vegetarian'' --- does not relate to singing at all. However, the \grover{} model produces singing-related content anyway (Bessie Smith was a singer).

\section{Reproducibility Details} 

This paper relies on the existing \fever{}, \scifact{}, and \covidfact{} datasets, which are publicly available. To test our method, we use the same evaluation metrics proposed by the dataset authors: label accuracy for \fever{} \citep{Thorne18Fever}, the sentence selection, sentence label, abstract label, and abstract rationalized metrics for \scifact{} \citep{wadden-etal-2020-fact}, and the Top-1 and Top-5 label accuracy for \covidfact{} \citep{Saakyan2021COVIDFactFE}. We also introduce our own datasets of adversarial evidence generated by \grover{} and PEGASUS \citep{zhang2019pegasus}. They will be made publicly available with a license that allows for research use.  
For computational experiments in this paper, the main source code is available at: \\

\url{https://github.com/Yibing-Du/adversarial-factcheck} \\

\noindent These experiments were run on an NVIDIA Quadro RTX 8000 GPU. Because we do not train these models from scratch, but instead use existing released models, we only run each evaluation once since the result is deterministic. Consequently, there are no hyperparameters to tune. We use the default hyperparameters provided with the codebases of the models evaluated.